%% file: actxFactor.tex
\documentclass[10pt,twocolumn,letterpaper]{article}

\usepackage{iccv}
\usepackage{times}
\usepackage{epsfig}
\usepackage{graphicx}
\usepackage{amsmath, amssymb}
\usepackage{color}
\usepackage[numbers,sort,compress]{natbib}
\usepackage{bbm}
\usepackage[pagebackref=true,breaklinks=true,letterpaper=true,colorlinks,bookmarks=false]{hyperref}
\usepackage{epsfig}
\usepackage{enumerate,booktabs}
\usepackage{multirow}
\usepackage{subcaption}
\usepackage[pagebackref=true,breaklinks=true,letterpaper=true,colorlinks,bookmarks=false]{hyperref}

\iccvfinalcopy % *** Uncomment this line for the final submission

\def\iccvPaperID{3149} % *** Enter the ICCV Paper ID here

\DeclareMathOperator*{\Norm}{AttNorm}
\newcommand{\AttNorm}[1]{\raisebox{0.5ex}{\scalebox{0.8}{$\displaystyle \Norm_{#1}\;$}}}
\DeclareMathOperator*{\Percent}{Percent}
\newcommand{\AttPercent}[1]{\raisebox{0.5ex}{\scalebox{0.8}{$\displaystyle \Percent_{#1}\;$}}}

\DeclareMathOperator*{\sigmoid}{sigmoid}
\DeclareMathOperator*{\softmax}{softmax}

\ificcvfinal\fi
\begin{document}
\input{definitions}

%%%%%%%%% TITLE
\title{Attentive Action and Context Factorization}

\author{
Yang Wang$^1$, ~Vinh Tran$^1$, ~Gedas Bertasius$^2$, ~Lorenzo Torresani$^3$, ~Minh Hoai$^1$ \\
$^1$Stony Brook University,~~$^2$University of Pennsylvania,~~$^3$Dartmouth College
%First Author\\
%Institution1\\
%Institution1 address\\
%{\tt\small firstauthor@i1.org}
% For a paper whose authors are all at the same institution,
% omit the following lines up until the closing ``}''.
% Additional authors and addresses can be added with ``\and'',
% just like the second author.
% To save space, use either the email address or home page, not both
\and
%Second Author\\
%Institution2\\
%First line of institution2 address\\
%{\tt\small secondauthor@i2.org}
}

\maketitle
\thispagestyle{empty}

%%%%%%%%% ABSTRACT
\begin{abstract}

    We propose a method for human action recognition, one that can localize the spatiotemporal regions that `define' the actions. This is a challenging task due to the subtlety of human actions in video and the co-occurrence of contextual elements. To address this challenge, we utilize conjugate samples of human actions, which are video clips that are contextually similar to human action samples but do not contain the action. We introduce a novel attentional mechanism that can spatially and temporally separate human actions from the co-occurring contextual factors. The separation of the action and context factors is weakly supervised, eliminating the need for laboriously detailed annotation of these two factors in training samples. Our method can be used to build human action classifiers with higher accuracy and better interpretability. Experiments on several human action recognition datasets demonstrate the quantitative and qualitative benefits of our approach.
   
\end{abstract}

%%%%%%%%% BODY TEXT
\section{Introduction}

Our work is concerned with human actions in video, and we consider the task of localizing the spatiotemporal regions of a video that define a human action. This task is useful for understanding human actions and for developing computational models of human actions with better accuracy and higher interpretability. This task, however, is very challenging due to the subtlety of human actions and the co-occurrence of other contextual elements in the video. Finding the constituent elements of a human action requires more than grounding the decisions of a human action classifier, because the decisions of a human action classifier might be partially based on some contextual cues such as the background scene and the camera motion.

To distinguish the spatiotemporal regions associated to actions from those corresponding to context, in this work we leverage conjugate samples of human actions, which are video clips that are contextually similar to human action samples, but do not contain the actions. Conjugate samples were used in the Action-Context Factorization (ACF) framework~\cite{m_Wang-Hoai-CVPR18} for training a human action classifier that can explicitly factorize human actions from the co-occurring context. However, while in ACF the objective was to learn factorized feature extraction networks for better classification performance adaptability, our goal here is to perform visual grounding of the two factors by automatically identifying the voxels associated to action and those corresponding to context.

\begin{figure}[t]
\begin{center}
\includegraphics[width=0.9\linewidth]{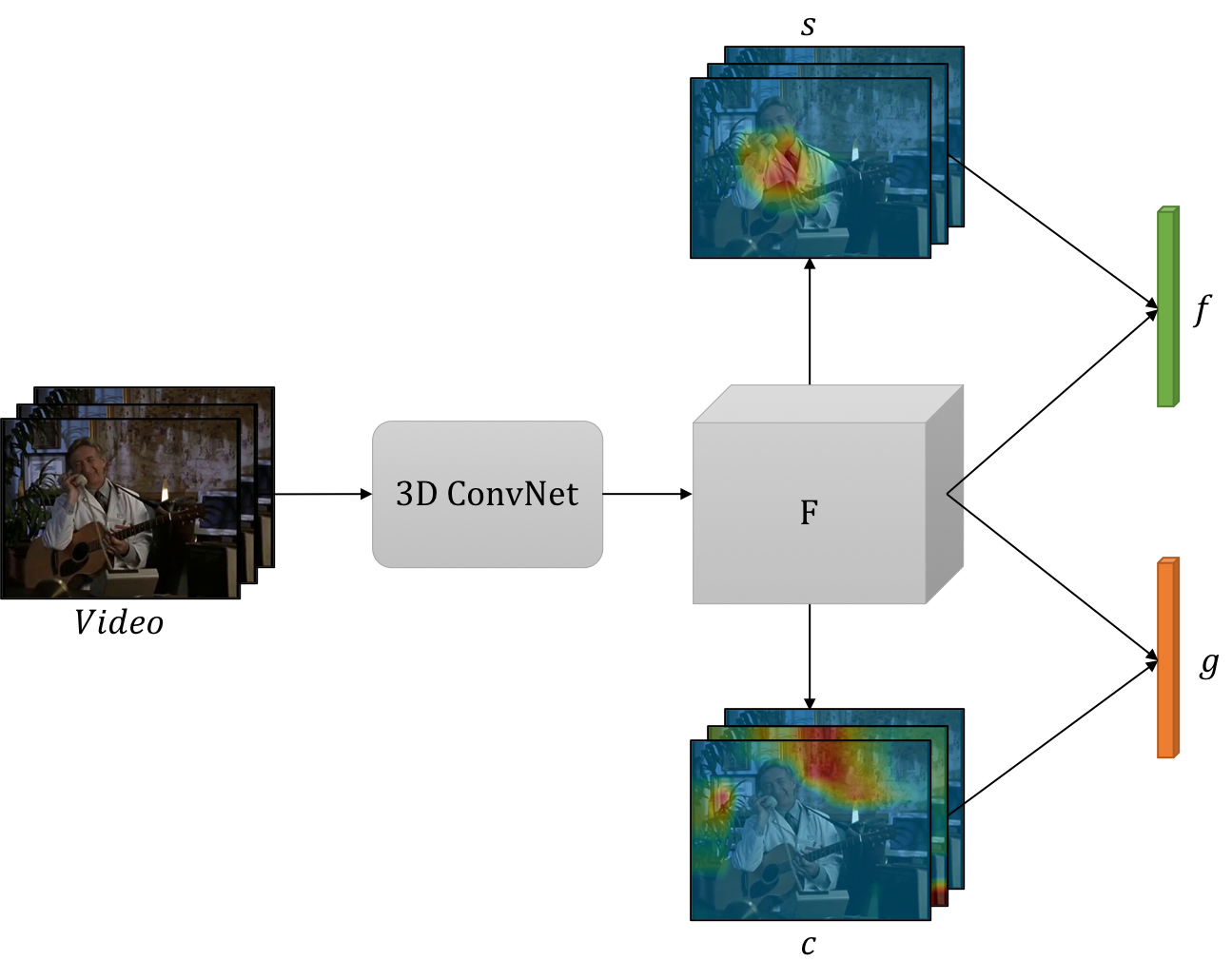}
\end{center}
   \vskip -0.2in
   \caption{{\bf Attentive Action and Context Factorization.} Given an input video clip, deep-learning feature maps are extracted and used to predict  two spatiotemporal attention maps, one for action and one for context. The feature vector representations for action and context will then be computed based on the attention maps.}
\label{fig: attentive_factorization}
\end{figure}

In this paper, we propose a novel attentional mechanism that can spatially and temporally  separate human actions from the co-occurring factors. 
Given an input video, this attentional mechanism will compute two spatiotemporal attention maps, one for the action components and one for the context components, as illustrated in Figure~\ref{fig: attentive_factorization}. Using these attention maps, we can identify and separate action pixels from context pixels. This allows us to selectively pool information from relevant regions of a video to compute feature vectors for the action and context parts of the video. This leads to an improved classifier with higher accuracy and better interpretability.

\section{Related Work}

\subsection{Visual grounding}

Given a query phrase or a referring expression, visual grounding requires a model to specify a region within the image or the video that corresponds to the query input. It is inspired by the top-down influences on selective attention in the human visual system (see~\cite{baluch2011mechanisms} for a review). Various methods~\cite{simonyan2014deepinside, zeiler2014visualizing, cao2015look, zhou2016learning, zhang2018top} have been proposed for grounding a CNN's prediction for images. However, visual grounding for video data or temporal network architectures is much less explored. 
\citet{karpathy2016visualizing} visualized LSTM cells that keep track of long-range dependencies in a character-based model.
\citet{selvaraju2017grad} presented qualitative results on grounding image captioning and visual question answering using an RNN. 
\citet{bargal2018excitation} proposed a top-down attention mechanism for CNN-RNN models to produce spatiotemporal saliency maps that can be used for action/caption localization in videos.

\subsection{Disentanglement in image/video synthesis}

In recent years, much effort~\cite{mathieu2016disentangling, denton2017unsupervised, lample2017fader, shu2017neural, shu2018deforming} has been spent on the task of learning explicitly disentangled representations and subsequently leveraging them to control image or video synthesis process. Early approaches used bilinear model~\cite{tenenbaum2000separating} and encoder-decoder network~\cite{huang2007unsupervised} for separating content and style for images such as faces and text in various fonts. 
The disentanglement of content and style was further explored in ~\cite{dosovitskiy2015learning, kulkarni2015deepinverse} for computer graphics applications. 
More recently, deep learning frameworks based on Variational Auto Encoders~\cite{Kingma-Welling-ICLR14, Mescheder2017ICML} and Generative Adversarial Networks~\cite{Goodfellow-etal-NIPS14, Radford-etal-arXiv15, chen2016infogan} have become more popular because they are more powerful than classical methods at encoding and synthesizing images and videos. Besides the disentanglement of content and style, some research~\cite{hinton2011transforming, shu2018deforming} has also been conducted on separating the feature representation into translation-related and translation-invariant factors.

\section{Attentive Factorization} 

\subsection{Framework Overview \label{sec:acframework}}

\begin{figure}[t]
\begin{center}
\includegraphics[width=0.9\linewidth]{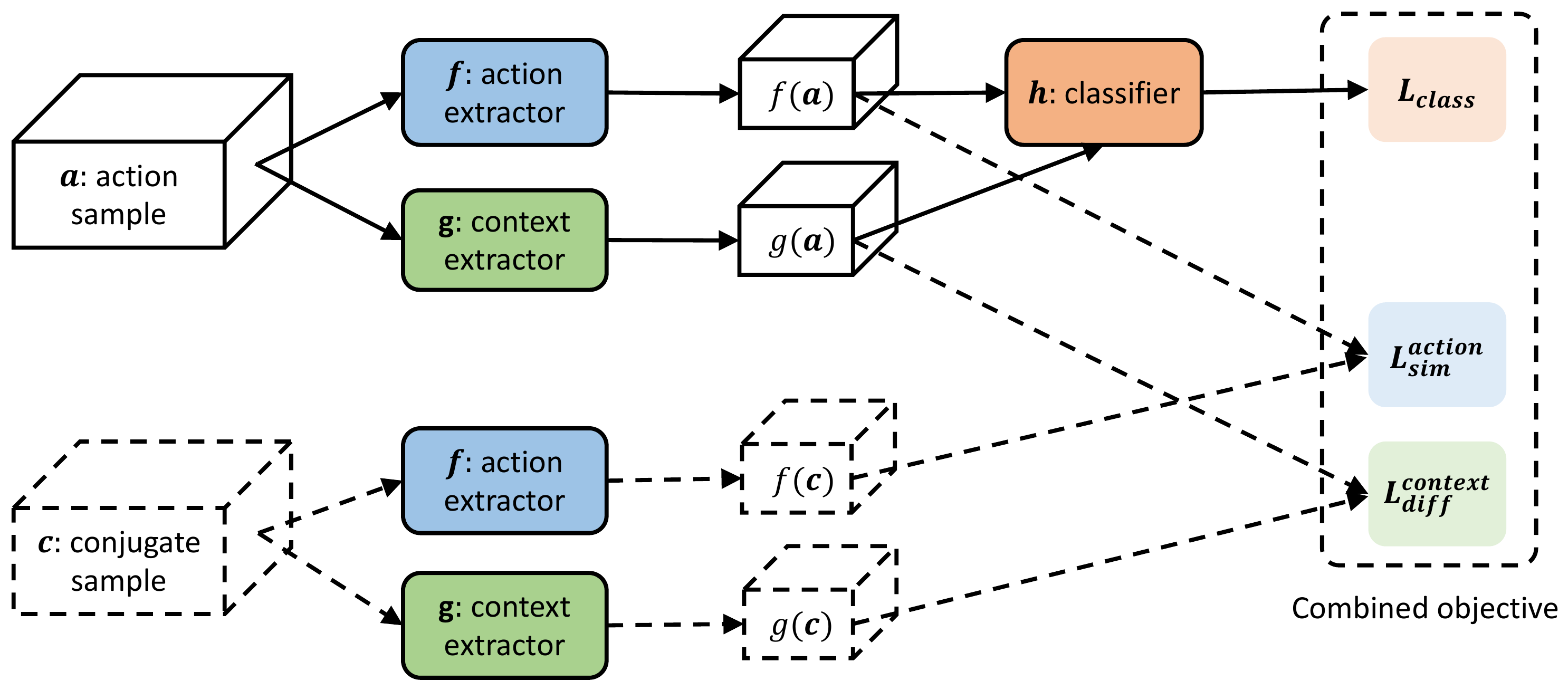}
\end{center}
\vskip -0.1in 
   \caption{{\bf A framework for attentive action and context factorization.} Attentive feature extraction and classifier learning are two stages of a joint learning framework. Functions $f$ and $g$ are for extracting the action and context feature maps respectively, using spatiotemporal attention maps. The extracted feature maps are subsequently fed into the classifier $h$. The objective is to minimize the classification loss and the similarity between two action feature maps $f(a)$ and $f(c)$, while maximizing the similarity between two context feature maps $g(a)$ and $g(c)$.}
\label{fig: loss_function}
\end{figure}

We propose to train a human action classifier with both video samples of human actions and conjugate samples. The training data is $\{(l_i, \a_i, \c_i)\}_{i=1}^{n}$, where $l_i$ is an action label, $\a_i$ is the video sample for action $l_i$, and $\c_i$ is a conjugate sample for $\a_i$. $\c_i$ is contextually similar to $\a_i$ but it does not contain the action $l_i$. We will seek two feature extractors $f$ and $g$ for extracting the feature maps to represent action and context components respectively. Since $\c_i$ does not contain the same action as $\a_i$, it is desirable that $f(\a_i)$ and $f(\c_i)$ are different. On the other hand, $\c_i$ shares similar context with $\a_i$, so $g(\a_i)$ should be similar to $g(\c_i)$. Based on this observation, we will learn the action and context extractors, together with an action classifier $h$, by minimizing the following loss function: 
\begin{align}	
	&\sum_{i=1}^n   \mL^{action}_{sim}(f(\a_i), f(\c_i)) + \sum_{i=1}^n\mL^{context}_{diff}(g(\a_i), g(\c_i)) \nonumber  \\
	&+ \sum_{i=1}^n \mL_{class}(h(f(\a_i), g(\a_i)), y_i). \label{eqn:acf}
\end{align}
Here, $\mL^{action}_{sim}$ is a loss function that penalizes the similarity between two action feature maps; $\mL^{context}_{diff}$ is a loss function that penalizes the difference between two context feature maps; and $\mL_{class}$ is a classification loss function that penalizes the difference between the output of the classifier $h$ (based on both the action and context features of the action sample $\a_i$) and the ground truth label $y_i$.

By optimizing Eq.~(\ref{eqn:acf}), we can learn the action and context feature extractors $f(\v)$ and $g(\v)$ for an input video $\v$. Our goal, however, is to perform visual grounding of the two feature representations to identify the voxels of $\v$ that contribute to these features, which is different from the goal of learning factorized feature embeddings in the ACF framework~\cite{m_Wang-Hoai-CVPR18}.
We therefore propose to incorporate an attentional mechanism in the parameterization of $f(\v)$ and $g(\v)$ as follows. We assume there is a feature extraction network and a video can be represented by a 4D feature map. Given an input video $\v$, many existing pretrained ConvNet models can be used to extract convolutional feature maps $F\in \Re^{T \times H \times W \times D}$ that span both in the spatial and temporal dimensions. Due to the use of strided convolutional or pooling layers within the applied ConvNet, the dimensions of the obtained feature maps may be different from the sizes of the original video. Usually, $T$, $H$, and~$W$  can be obtained by dividing the length, the height, and the width of the video $\v$ by their effective convolutional strides respectively. $D$ is the number of output channels of the applied ConvNet. Given the feature map $F$, we will compute two attention maps  $S$ and $C \in [0,1]^{T\times H \times W}$ for the spatiotemporal distribution of the action and the context elements of the video respectively, as illustrated in Figure~\ref{fig: attentive_factorization}. $s_i$ is the probability that the voxel $i$ of the feature map $F$ corresponds to an action component, while $c_i$ is the probability that the voxel $i$ is a context element. Finally, the action extractor $f(\v)$ can be defined as the function of the map $F$ and the action map $S$, while the context extractor is the function of $F$ and the context map $C$. Further details will be provided in subsequent subsections. 

One technical challenge we address is the  design of proper loss functions for $\mL_{class}$, $\mL^{action}_{sim}$, and $\mL_{diff}^{context}$. While the first loss can be defined based on any standard classification loss such as the cross entropy loss, the two latter losses $\mL^{action}_{sim}$ and $\mL_{diff}^{context}$ cannot be chosen arbitrarily. Although it seems intuitive to define these loss functions based on the  $L_2$ or cosine distances between two feature vectors/maps, this choice would yield poor performance. To see the problem, consider an illustrating example where $\mL^{context}_{diff}$ is defined based on the cosine distance: $\mL^{context}_{diff}(g(\a_i), g(\c_i)) = 1 - \cos(g(\a_i), g(\c_i))$. This loss function assumes that $g(\a_i)$ and $g(\c_i)$ have the same cardinalities, and they are aligned voxel to voxel. But video is a dynamic environment and there might be positional shifts between the contextual elements in the action sample $\a_i$ and those in the conjugate sample $\c_i$. Thus, it might be too forceful to require $g(\a_i)$ and $g(\c_i)$ to be perfectly aligned, unless $g(\a_i)$ and $g(\c_i)$ are the globally-pooled feature representations. However, globally-pooled  representations do not preserve locality, while locality is very important for visual grounding.

In the next subsection, we will describe the loss functions for measuring the similarity and difference between weakly-aligned concepts. After that, we will describe the attentional mechanism for visual grounding. The attention maps, the feature extractors, and the action classifier will be jointly learned end-to-end as illustrated in Figure~\ref{fig: loss_function}.

\subsection{Comparing weakly-aligned feature maps}

Let $\X = F(\a)$ be the 4D feature map for an action sample and~$\Z = F(\c)$ the feature map for the corresponding conjugate sample. 
For brevity of the below presentation, let us reshape~$\X$ and~$\Z$ into 2D matrixes: $\X = \big[\x_1, \cdots, \x_N\big] \in \Re^{D \times N}$ and~$\Z = \big[\z_1, \cdots, \z_M\big] \in \Re^{D \times M}$, where $N$ and $M$ are the number of voxels in $\X$ and $\Z$ respectively. The temporal lengths of an action sample and conjugate sample might be different so $N=T_a \times H \times W$ might be different from $M=T_c \times H \times W$. 

In general, it is not a good idea to measure the element-wise similarities and differences between $\X$ and $\Z$ due to the size difference and the positional shift of action and context elements. To address this problem, we propose the following mechanism. Consider a feature vector $\x_i$ at location $i$ of the feature map for the action sample. By definition, if $\x_i$ corresponds to an action element, it should be different from $\z_j$ for all $j$. On the other hand, if $\x_i$ corresponds to a context voxel, then there must be $j$ in $\Z$ such that $\z_j$ is similar to $\x_i$. We propose to define a quantity to measure the amount of $\x_i$ in $\Z$ as follows: 
\begin{align}
\eta(\x_i, \Z) = \sum_j \frac{\exp(\gamma \hat{\x}_i^T\hat{\z}_j)}{\sum_{j'}\exp(\gamma \hat{\x}_i^T\hat{\z}_{j'})}\z_j.	 \label{equ: attentive_pool}
\end{align}
where $\hat{\x}_i$ and $\hat{\z}_j$ are unit vectors. Equivalently, $\eta(\x_i, \Z)$ can be expressed in terms of matrix-vector multiplication  $\eta(\x_i, \Z) = \Z\softmax(\gamma\hat{\Z}^T\hat{\x}_i)$. Similarly, we can compute a vector $\eta(\x_i, \X)$ to represent the amount $\x_i$ in $\X$. $\eta(\x_i, \X)$ and $\eta(\x_i, \Z)$ should be different or similar depending on whether $\x_i$ corresponds to  an action or context voxel.  

\begin{align}
&\mL^{action}_{sim}(\X, \Z) = \frac{\sum_i s_i \times \textrm{sim}(\eta(\x_i, \X),~\eta(\x_i, \Z))}{\sum_i s_i}, \\
&\mL_{diff}^{context}(\X, \Z) = \frac{\sum_i c_i \times  \textrm{diff}(\eta(\x_i, \X),~\eta(\x_i, \Z))}{\sum_i c_i}.
\label{equ:unaligned-case}	
\end{align}
In the above, $\textrm{sim}$ and $\textrm{diff}$ are two functions that measure the similarity and difference between two vectors based on their cosine value. Specifically in our experiments, we set $\textrm{sim}(\x, \z) = \max \big\{ 0, ~\cos \big\langle \x, \z \big\rangle - \xi \big\}$ and $\textrm{diff}(\x, \z) = 1 - cos\big\langle \x, \z \big\rangle$. 

The attentional pooling mechanism in Equation \ref{equ: attentive_pool} is feature similarity-based. It resembles methods such as non-local network~\cite{wang2018nonlocal}, double-attention network~\cite{chen2018double}, and transformer network~\cite{vaswani2017transformer}. The key difference between our approach and the previous methods is that we aim to compare features across different videos, while the prior approaches aim to reweigh features within a single example.

\subsection{Attentive Action \& Context Factorization} 

As mentioned above, a video is represented by a 4D convolutional feature map $F\in \Re^{T\times H\times W \times D}$, and we will compute two attention maps  $S$ and $C \in [0,1]^{T\times H \times W}$ for the spatiotemporal distribution of the action and the context elements of the video. In this section, we describe the parameterization of $S$ and $C$ as functions of $F$.

\begin{figure}[t]
\begin{center}
\includegraphics[width=\linewidth]{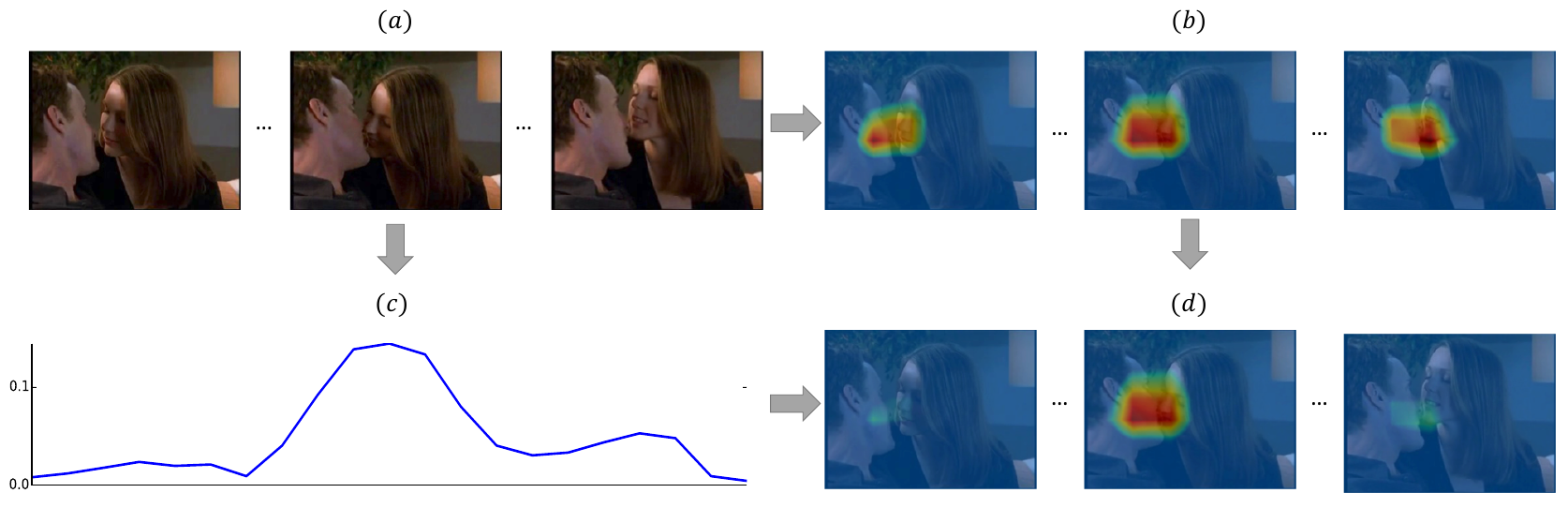}
\end{center}
\vskip -0.1in 
   \caption{{\bf Visualizing the spatiotemporal attention map $S^{att}$.} This is an example of the action Kiss. (a): Input video. (b): Spatial attention maps if normalized within each frame using a softmax function. (c): Temporal attention map if normalized across time steps. (d): Spatiotemporal attention map $S^{att}$ normalized among the entire video volume; this is a non-negative unit-sum 3D array. 
   \label{fig:attentive_feat_comb}}
\end{figure}

We propose to define the spatiotemporal attention maps $S$ and $C$ based on one action map $S^{act} \in [0,1]^{T \times H \times W}$ and one attention map $S^{att} \in [0,1]^{T \times H \times W}$ as below:
\begin{equation}
\begin{split}
S = S^{act} \odot S^{att}, \; C = 1 - S^{act}.
\end{split}
\end{equation}
Operator $\odot$ denotes element-wise multiplication. 
The action map $S^{act}$ is the output of a per-location sigmoid function. Its values represent the probability of each location being an action component. Conversely, $1 - S^{act}$ represents the probability of each location being a contextual element that is common in both the action sample and the conjugate sample. We use $S^{att}$ to further refine the action attention map $S$, because not all action elements are equal for the task of action recognition. $S^{att}$ is a unit-sum tensor that allocates the percentage of attention for each location. It is produced by a softmax function instead of a sigmoid function, and should attend to more discriminative action components, as showcased in Figure \ref{fig:attentive_feat_comb}.

With the convolutional feature map $F \in \Re^{T\times H\times W \times D}$ that represents the video, and two spatiotemporal attention maps $S$ and $C \in [0,1]^{T\times H \times W}$ respectively for the action and the context elements, we propose to compute the action feature and the context feature as follow.
\begin{equation}
\begin{split}
F_s &= \frac{1}{\sum s_i} \sum_i s_i \cdot \theta(F)_i,  \\
F_c &= \frac{1}{\sum c_i} \sum_i c_i \cdot \theta(F)_i.
\end{split}
\label{equ: factorize}
\end{equation}
$F_s$ and $F_c$ are the action and the context feature representations for input video $\v$. $\theta$ is a learnable transformation that is applied to the original feature maps. In this work, we define $\theta(F) = relu(F + conv3d(F))$, which includes a residual connection to aid the optimization process~\cite{He-et-al-CVPR16}. Other transformation functions are also plausible. Finally, we concatenate $F_s$ and $F_c$ and apply a fully-connected layer to obtain the class confidence scores.

\subsection{Implementation Details}

We now provide more details on the parameterization of $S^{act}$ and $S^{att}$ in Equation \ref{equ: intermediate_attentive_map}.
We first apply two separate 3D convolutional layers $\phi$ and $\psi$ on the extracted feature map $F$ to compute the action confidence map $\phi(F)$ and the attention confidence  map $\psi(F)$. These are unnormalized confidence scores, and their range highly depends on the values of the convolution kernel weights of $\phi$ and $\psi$. To ensure consistent value ranges across different videos and time steps, we apply a normalization step, called $\AttNorm{dims}$. This function can be applied on a multidimensional input tensor; it will subtract the average value from the input tensor and then divide the result by the standard deviation. Both the average value and the standard deviation are computed along the specified dimensions $dims$. This $\AttNorm{dims}$ function is similar to a Batch-Norm layer with a key difference. The proposed normalization is performed among different spatial locations at different time steps, instead of different samples in a single batch. We also use a $\AttPercent{\rho}$ module to control the percentage of positive values to be $\rho \in [0, 100]$ in each confidence map. This can be done by subtracting the $\rho$-percentile value from every element of the confidence map. By explicitly controlling the percentage of positive values before the sigmoid function, as used in Eq.~(\ref{equ: intermediate_attentive_map}), we can regulate the percentage of action components in $S^{act}$ within the entire video. Both $\AttNorm{dims}$ and $\AttPercent{\rho}$ are parameter-free. They enable efficient and stable network training in our experiments. After the normalization of the confidence scores, we apply the relaxed sigmoid or softmax function to obtain the intermediate attentive distribution maps:
\begin{equation}
\begin{split}
&S^{act} = \sigmoid(\alpha \cdot   \Percent_{\rho} \Norm_{t, h, w} \phi(F)),\\
&S^{att} = \softmax_{t, h, w}(\beta \cdot  \Norm_{t, h, w} \psi(F)).
\end{split}
\label{equ: intermediate_attentive_map}
\end{equation}
In the above, $\rho$, $\alpha$, and $\beta$ are tunable hyper parameters. $\rho$ controls the percentage of action component within each video. $\alpha$ controls the sharpness of the boundary between action and context. When $\alpha$ is 0, $S^{act}$ is $0.5$ everywhere, and every location contains equal amounts of action and context values. If $\alpha = \infty$, $\rho$ percent of the values in $S^{act}$ are 1's, and the rest are 0's. That entails a hard separation of action and context spatiotemporally. $\beta$ can be used to control the balance between average pooling and max pooling for the attention map. When $\beta$ is 0, $S^{att}$ has uniform values, and this is equivalent to average pooling. If $\beta=\infty$, only one location in the video has the attention weight of 1, while the attention weights of other locations are 0; this is equivalent to max pooling. 

The learnable parameters of the proposed action and context attentive mechanism are the kernel parameters of the functions $\phi$, $\psi$ and $\theta$. The functions $\AttNorm{dims}$, $\AttPercent{\rho}$, $\sigmoid$ and $\softmax$ are all parameter-free. $\rho$, $\alpha$ and $\beta$ are hyper-parameters which can be tuned using validation data; in our experiments, a good default starting point is $\rho=30.0, \alpha=\beta=1$.

\section{Experiments}

We perform experiments on several challenging action recognition datasets: ActionThread~\cite{Hoai-Zisserman-ACCV14}, Hollywood2 \cite{Marszalek-et-al-CVPR09}, HACS~\cite{zhao2019hacs}, and Pascal VOC~\cite{Everingham-et-al-VOC12}. We consider both spatiotemporal and spatial data, and we also consider imperfect conjugate samples. 

\subsection{Separating action and context in video}

\subsubsection{Dataset}
Our experiments in this section are performed on the ActionThread dataset~\cite{Hoai-Zisserman-ACCV14}. This dataset contains not only the video clips that include human actions but also the sequences right before and after the actions. These sequences are good candidates for conjugate samples of human actions. The human action samples in ActionThread were automatically located and extracted using script mining in 15 different TV series. ActionThread has 13 different actions and 3035 video clips, which are split into disjoint train and test subsets~\citep{Hoai-Zisserman-ACCV14}. We consider the pre- and post-action sequences as the source for mining conjugate samples, and make sure each pair of action and conjugate samples are extracted from the same video thread.

\subsubsection{Feature extraction} 

We use the pre-trained two-stream I3D ConvNet~\cite{carreira2017i3d} for feature extraction. Each video is represented as two 4D feature maps $F_{rgb}, F_{flow} \in \Re^{T\times 8 \times 11 \times 1024}$. $T$ is the number of video frames divided by 8. The proposed attentive factorization network can take input feature maps with different temporal lengths. However, temporal cropping with a fixed length is still necessary during training because we need to put feature maps into mini-batches. We use a batch size of 36 videos and the temporal cropping length is set to 10, equivalent to 80 frames or 3.2 seconds. At test time, temporal cropping is unnecessary. The entire feature maps are fed into the network to perform attentive action and context factorization as well as action recognition.

\myheading{More details.} We use the I3D model~\cite{carreira2017i3d} that was trained on both ImageNet~\cite{Russakovsky-etal-IJCV15} and Kinetics~\cite{Kay-etal-arxiv17} datasets. We choose I3D ConvNet for feature extraction because it is the current state-of-the-art method for human action recognition. Specifically, we use the output of ``Mixed\_5c'', the last convolutional layer before global average pooling, as our convolutional feature map $F$. The effective accumulated convolutional stride at this layer are 32, 32, 8 for vertical, horizontal, temporal dimensions respectively. We always resize the input video frames to have height $\mathcal{H} = 256$ and width $\mathcal{W} = 352$, thus the output feature map has height $H=\mathcal{H}/32=8$ and width $W=\mathcal{W}/32=11$. For an input video clip that spans $\mathcal{T} = 128$ frames, the output feature map would have the temporal length $T=\mathcal{T}/8=16$.

\subsubsection{Action recognition performance} 
Table \ref{table: ActionThread} compares the action recognition performance of several methods on the ActionThread dataset. {\it  I3D} refers to the baseline approach~\cite{carreira2017i3d}, where the I3D feature vectors are averaged over the entire video. {\it ActX-I3D} is the proposed approach where attentive action and context factorization and human action classification are jointly optimized. As can be seen from Table \ref{table: ActionThread}, the proposed ActX-I3D approach significantly outperforms I3D (a direct baseline method without an attentional mechanism). The improvement brought by ActX-I3D, compared to the baseline I3D, is most evident on the spatial stream where the RGB images are used for extracting features.

We also compare our method with alternative approaches for estimating the attention maps and use them for weighting the I3D feature maps. 
{\it ACF-I3D} refers to the approach of~\cite{m_Wang-Hoai-CVPR18} where $\mL_{sim}^{action}$ and $\mL_{diff}^{context}$ are defined based on cosine distance between globally pooled feature vectors. 
{\it Attentional-I3D} refers to a state-of-the-art attentional pooling method~\cite{Girdhar_17b_AttentionalPoolingAction} where both the top-down (category-specific) and the bottom-up (category-agnostic) attention maps are learned for weighting the predicted action scores. 
 {\it PoseMask-I3D} refers to a method where  human poses are used to obtain an attentive action map that focuses on regions of the human body joints. We run the Convolutional Pose Machine algorithm~\cite{Wei-et-al-CVPR16} from the OpenPose software library software to detect human pose keypoints within each video frame. The detected body keypoints include ears, eyes, and mouth regions, as well as shoulders, hips, limbs, wrists, and ankles. We convert these keypoint heat maps into binary masks and use them for weighted pooling of the I3D features. We hope the pose mask can help focusing the classifier's attention on the human regions and achieving better action recognition results. However, the performance of PoseMask-I3D is worse than the average-pooled I3D features, indicating the importance of non-human regions. Because of this, we also consider {\it $k$-Region-I3D} where object region proposals are used for attentive attribution. We use the Detectron software library~\cite{Detectron2018} to obtain the top $k$ object regions predicted by the Region Proposal Network of a Mask RCNN. We use ROI-align module to obtain the feature representation for each region and average pool them into a single feature vector. We experimented $k=10$ and $k=50$, both yielding an action recognition performance that is slightly better than the average pooling method (I3D). However, these methods are still outperformed by the proposed method ActX-I3D.
We also perform an ablation study where $S^{att}$ is not used (i.e., $S = S^{act}$). The performance is not as good as when $S^{att}$ is used to refine the action attention map. The amounts of parameters added by these attentional mechanisms are negligible compared to that of the I3D backbone network.

\setlength{\tabcolsep}{6pt}
\begin{table}[t]
\begin{center}
\begin{tabular}{lrcrr}
\toprule
\multirow{2}{*}{Method} & \multicolumn{2}{c}{RGB Only} & \multicolumn{2}{c}{RGB + Flow} \\
\cmidrule(lr){2-3} \cmidrule(lr){4-5} 
& \#param & mAP & \#param & mAP \\
\midrule
I3D~\cite{carreira2017i3d}  & 13.4M & 46.7 & 28.8M  & 55.9 \\
PoseMask-I3D & n/a & 46.0 & n/a & 53.2 \\
10Region-I3D & n/a & 50.1 & n/a & 57.6 \\
50Region-I3D & n/a & 49.6 & n/a & 56.3 \\
Attentional-I3D ~\cite{Girdhar_17b_AttentionalPoolingAction} & 13.4M & 53.0 & 28.8M & 61.8 \\
ACF-I3D~\cite{m_Wang-Hoai-CVPR18} & 13.4M & 52.9 & 28.8M & 60.4 \\
ActX-I3D [\small{w/o $S^{att}$}]  & 13.4M & 51.2 & 28.8M & 58.8 \\
ActX-I3D [\small{Proposed}]  & 13.4M & \textbf{55.4} & 28.8M & \textbf{63.1} \\
\bottomrule
\end{tabular}
%\vskip 0.1in
\caption{{\bf Action recognition results on the ActionThread dataset}. The table shows average precision values (higher is better). All methods are based on the same feature maps produced by the I3D backbone. I3D~\cite{carreira2017i3d} uses global average pooling. All other methods use attention for weighted pooling of features. The proposed method ActX-I3D achieves the best performance. We also perform an ablation study where $S^{att}$ is not used. The performance is not as good as when $S^{att}$ is used to refine the action attention map. The amounts of parameters added by these attentional mechanisms are negligible compared to that of the I3D backbone. }
\label{table: ActionThread}
\vspace*{-0.15in}
\end{center}
\end{table}
\setlength{\tabcolsep}{1.4pt}

%\begin{figure*}[t]
%\begin{center}
%\includegraphics[width=0.9\linewidth]{images/at_results.pdf}
%\end{center}
%\vskip -0.1in 
%   \caption{{\bf Examples of the spatiotemporal attention maps for action and context.} $S^{frame}$ is the attention along the temporal direction. $S$ and $C$: spatiotemporal attention maps of action and context respectively. The action class is: Eat. 
%\label{fig:at_results}}
%\end{figure*}

\subsubsection{Visualizing the action and context components}

\begin{figure}
\centering
\makebox[0.45\linewidth]{Action attention map} 
\makebox[0.45\linewidth]{Context attention map} 
\includegraphics[width=0.45\linewidth]{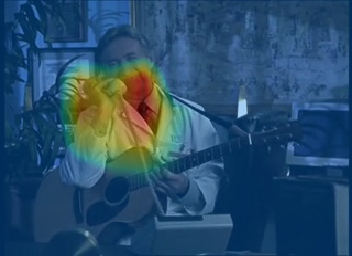}
\includegraphics[width=0.45\linewidth]{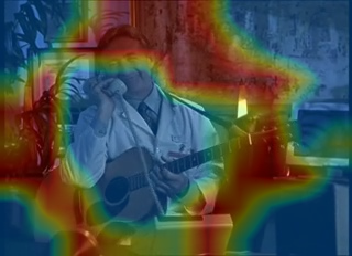}
\includegraphics[width=0.45\linewidth]{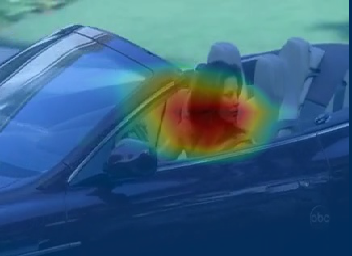}
\includegraphics[width=0.45\linewidth]{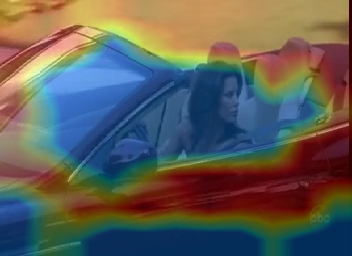}
\includegraphics[width=0.45\linewidth]{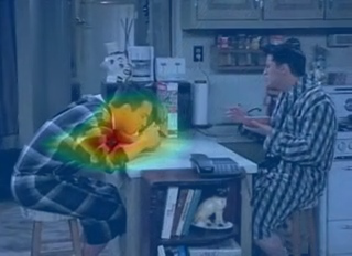}
\includegraphics[width=0.45\linewidth]{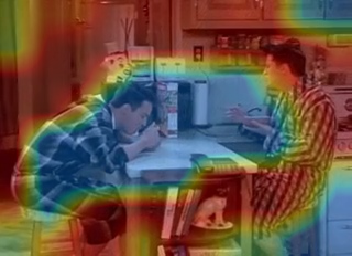}
\includegraphics[width=0.45\linewidth]{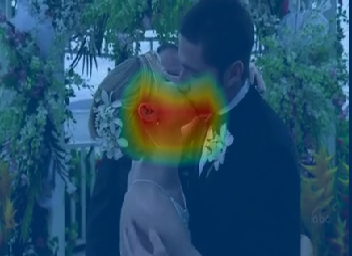}
\includegraphics[width=0.45\linewidth]{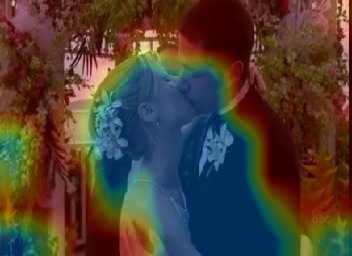}
\caption{{\bf Examples of attention maps for action and context.} From top to bottom, the actions are: AnswerPhone, DriveCar, Eat, and Kiss. The action attention maps have higher weights on humans, but not all humans receive the same attention, and not all parts of a human subject receive attention. The weights of the context maps are lower on the human subjects. The context maps have nonuniform distribution over background pixels. 
\label{fig:attmaps}}
\end{figure}

Fig.~\ref{fig:attmaps} shows some examples of action and context attention maps. To recognize a human action, it is important to attend to the human subjects, and this explains why the attention maps for action have higher weights on human subjects. However, not all humans in a video frame receive the same attention, and not all body parts of a human receive attention. On the contrary, the weights of the context maps are lower on the human subjects. The context maps emphasize the background regions with a nonuniform distribution.

\subsection{Imperfect conjugate samples}

Until now, we have assumed the availability of perfect conjugate samples that do not contain the actions in consideration. This assumption holds in general. For example, if a human action dataset has temporal annotations for the start and end times of the actions, we can consider the sequences right before and after the action boundaries as conjugate samples. Or in another scenario where a new action dataset is being collected, we can save the pre- and post-action sequences  for future mining of conjugate samples. However, if we are forced to work with a specific dataset such as Hollywood2, where the pre- and post-action sequences are not available, we can use ``imperfect'' conjugate samples instead. In this scenario, the conjugate samples are not guaranteed to exclude all action elements that are shared with the action samples. With this in mind, we investigate whether or not a black-and-white separation between conjugate and action samples with respect to the action elements are necessary.

\myheading{Experiments on Hollywood2.} We first perform experiments on the Hollywood2 dataset~\cite{Marszalek-et-al-CVPR09}, which contains 12 actions and 1,707 videos collected from 69 Hollywood movies. For training the proposed method on Hollywood2, we use a batch size of 36 videos, and the temporal cropping length for I3D feature maps is set to 6, equivalent to 48 frames or around 2 seconds. The source for mining conjugate samples is the sequences right before and after the temporally cropped action samples. In this way, the extracted action and conjugate samples share the same context, and the key difference between them is the dynamic content of the human action at different action stages.

Table \ref{table: Hollywood2} compares the action recognition performance of several methods on the Hollywood2 dataset. {\it ActX-I3D} is the proposed approach where attentive action and context factorization and human action classification are jointly optimized. It significantly outperforms the baseline {\it I3D}~\cite{carreira2017i3d} approach where the I3D feature vectors are averaged over the entire spatiotemporal domain. We also observe that the performance gaps between the proposed {\it ActX-I3D} method and other alternative attentive models on the Hollywood2 dataset are similar to that on the ActionThread dataset, presented in Table \ref{table: ActionThread}. More specifically, the proposed {\it ActX-I3D} significantly outperforms {\it Attentional-I3D}, where both the category-specific and agnostic attention maps are used, while {\it PoseMask-I3D, $k$-Region-I3D} achieve worse or similar action recognition performance than the average-pooled I3D features. 
The last two rows of Table \ref{table: Hollywood2} also compares the action recognition performance of the baseline I3D features and the proposed ActX-I3D features, when they are temporally compressed using Eigen Evolution Pooling~(EEP)~\cite{m_Wang-etal-FG18}. EEP uses a set of basis functions learned from data using PCA to encode the evolution of features over time. It provides an effective way to capture the long-term and complex dynamics of human actions in video. ActX-I3D still outperforms I3D when combined with EEP.

\setlength{\tabcolsep}{6pt}
\begin{table}[t]
\begin{center}
\begin{tabular}{lcccc}
\toprule
Method & RGB Only & RGB + Flow\\
\midrule
I3D~\cite{carreira2017i3d}  & 55.3 & 62.8 \\
PoseMask-I3D  & 49.4  & 60.0 \\
10Region-I3D  & 55.6  & 64.6 \\
50Region-I3D  & 55.9  & 64.8 \\
Attentional-I3D ~\cite{Girdhar_17b_AttentionalPoolingAction} & 61.3 & 69.8 \\
ACF-I3D~\cite{m_Wang-Hoai-CVPR18}  & 61.3 & 71.2 \\
ActX-I3D [Proposed]  & \textbf{62.5} & \textbf{73.2} \\
\midrule
EEP + I3D~\cite{m_Wang-etal-FG18}  & 73.5 & 81.0 \\
EEP + ActX-I3D [Proposed] & \textbf{76.0} & \textbf{82.0} \\
\bottomrule
\end{tabular}
%\vskip -0.1in
\caption{{\bf Action recognition results on the Hollywood2 dataset.} The table shows average precision values (higher is better). All methods are based on the same feature maps produced by the I3D ConvNet. The proposed ActX-I3D achieves better performance than other attention mechanisms, when used with or without Eigen Evolution Pooling~(EEP)\cite{m_Wang-etal-FG18}. We consider EEP because it has the current state-of-the-art performance on Hollywood2.}
\label{table: Hollywood2}
%\vspace*{-0.15in}
\end{center}
\end{table}
\setlength{\tabcolsep}{1.4pt}

\setlength{\tabcolsep}{6pt}
\begin{table}[t]
\begin{center}
\begin{tabular}{lcccc}
\toprule
Method & mAP & mAcc \\
\midrule
Backbone (3D-Res34-RGB)~\cite{hara3dcnns}  & 57.8 & 83.1 \\
Attention ~\cite{Girdhar_17b_AttentionalPoolingAction} & 58.6 & 84.5 \\
ACF~\cite{m_Wang-Hoai-CVPR18}  & 58.3 & 84.3 \\
ActX [Proposed]  & \textbf{59.1} & \textbf{85.3} \\
\bottomrule
\end{tabular}
%\vskip -0.1in
\caption{{\bf Action recognition results on the HACS-30 validation set.} The table shows mean average precision and mean class accuracy values (higher is better). All methods are based on the same feature maps produced by the backbone network (3D-Res34-RGB ~\cite{hara3dcnns}). The proposed ActX approach achieves better performance than other attention mechanisms.}
\label{table: HACS}
%\vspace*{-0.15in}
\end{center}
\end{table}
\setlength{\tabcolsep}{1.4pt}

\myheading{Experiments on HACS-30.} We also perform an experiment on the HACS dataset~\cite{zhao2019hacs}. This dataset provides both positive and negative samples for a list of 200 actions. Some of the negative samples and the positive ones are extracted from the same videos and they share very similar context. The key difference is whether an action of interest is observed or not. Thus the negative samples are a great source for mining conjugate samples. More precisely, given a positive action sample, we consider its top three nearest neighbors in the negative data pool as the conjugate samples. To compute the distance between videos, we use feature embeddings extracted by the 3D-Res34-RGB~\cite{hara3dcnns} network. 

Limited by computation resource, we use a subset of the HACS dataset, which we will refer to as HACS-30. This subset contains 30 positive action classes and one negative background class. The 30 actions are those with the least positive samples, which makes the recognition task more challenging. After conjugate mining, the training set of HACS-30 contains 11770 positive samples, and each positive sample has three corresponding conjugate samples. We evaluate the action recognition performance on the HACS-30 validation set, which contains 1186 samples for the 30 actions and 1712 samples for the negative background class. Given the imbalance between positive and negative data, we report the mean average precision (mAP) and the mean class accuracy (mAcc; obtained by averaging the per-class accuracies over 31 classes). The experiment results are reported in Table \ref{table: HACS}. We choose 3D-Res34-RGB as the backbone network because it outperforms I3D on HACS-30. As can be observed, the proposed ActX achieves better performance than other attention mechanisms. However, the performance improvement is less than that achieved on the ActionThread and Hollywood2 dataset. This might be due to the short duration of HACS video clips (all clips are only 2 seconds long) and the attention mechanism is less effective for short videos.

\subsection{Spatial action and context attention maps}

\begin{figure}[t]
\begin{center}
\includegraphics[width=\linewidth]{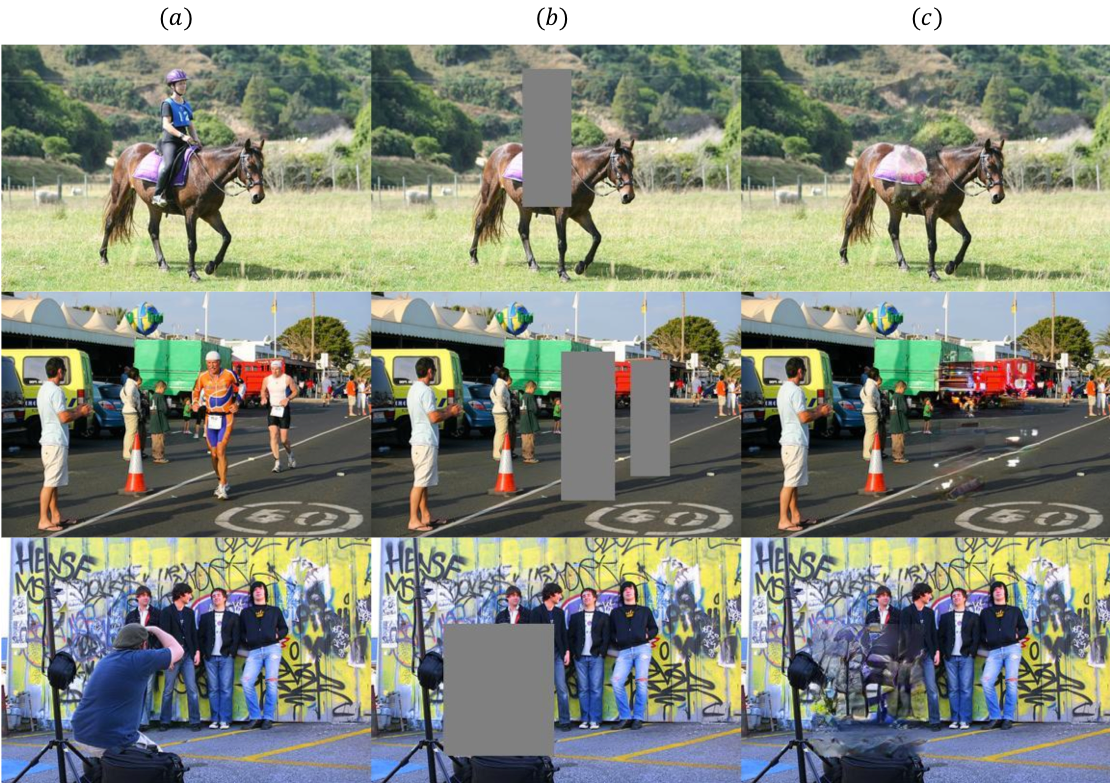}
\end{center}
\vskip -0.15in
   \caption{{\bf Action and conjugate samples for Pascal VOC2012 dataset.} From top to bottom: \textit{Riding Horse}, \textit{Running} and \textit{Taking Photo}. (a): action samples; (b) the corresponding bounding boxes for people performing the action in consideration; (c) conjugate samples obtained with image completion.}
\label{fig: inpaint}
\end{figure}

We propose a method to produce attention maps for spatiotemporal data. However, our method can also be used for spatial data. In this section, we perform experiments on still images, demonstrating the ability of our method for identifying action and context regions of an image. Specifically, we used the Pascal VOC2012 action dataset~\cite{Everingham-et-al-VOC12}. This dataset contains still images of 10 actions: Jumping, Phoning, Playing Instrument, Reading, Riding Bike, Riding Horse, Running, Taking Photo, Using Computer, and Walking. Some images in this dataset contain multiple people, and different people may perform different actions. Furthermore, a person may perform more than one action simultaneously, such as {\it Walking} and {\it Phoning}.

For this dataset, we consider an action sample to be the entire image without any bounding box information. We generate a conjugate sample for each image using the following steps. First, for the action being considered, we identify all the people performing this action in the image. This step will return a set of human bounding boxes for one or multiple people in the image if there are multiple people performing the same action. Second, we remove the pixels in identified bounding boxes of the previous step. Third, a pre-trained image completion network~\cite{IizukaSIGGRAPH2017, wang2018videoinp} is applied to fill in the missing regions with alternative content. This network is composed of one generator for image completion and two discriminators for the local and global context respectively in order to determine that the generated image be completed consistently. Finally, we perform a post-processing step by blending~\cite{Perez:2003:PIE:882262.882269, telea2004image} the filled regions with the surrounding pixels. Some action samples and the corresponding conjugate samples generated by the network are shown in Fig.~\ref{fig: inpaint}.

\begin{figure}
\centering
\makebox[0.45\linewidth]{Action attention map} 
\makebox[0.45\linewidth]{Context attention map} 
\includegraphics[width=0.45\linewidth]{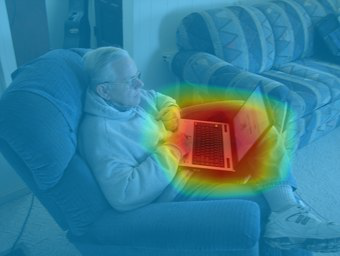}
\includegraphics[width=0.45\linewidth]{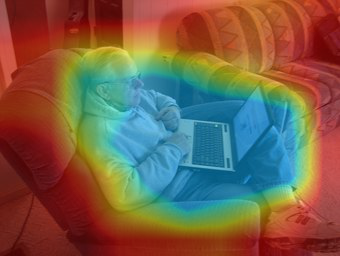}
\includegraphics[width=0.45\linewidth]{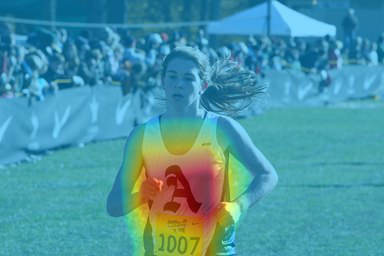}
\includegraphics[width=0.45\linewidth]{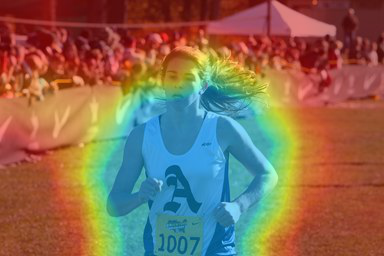}
\includegraphics[width=0.45\linewidth]{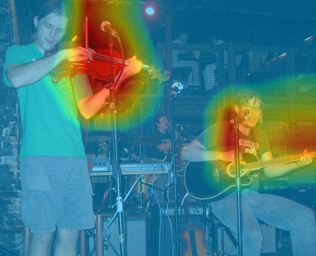}
\includegraphics[width=0.45\linewidth]{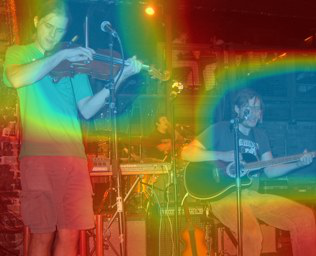}
\caption{{\bf Examples of attention maps for action and context on Pascal VOC2012 dataset.} From top to bottom, the actions are: \textit{Using Computer, Running}, and \textit{Playing Instrument}. Similar to video dataset, the action maps put more attention on the human object interaction, and the context maps focus more on the the surrounding area.
\label{fig:attmaps_voc}}
\end{figure}

We use a pre-trained DenseNet-161 model for feature extraction.  Each image is represented as a 3D features map $F\in \Re^{H \times W \times D}$. The sizes of images are different, so we resize the smaller dimension to $256$ before feeding to the network. During training, we extract random crop of size $8 \times 8$ on the feature map and train the network with the mini-batch size of $32$. At test time, the entire feature maps are fed into the network for attentive factorization and recognition.

The baseline model using DenseNet-161 features has a mean AP of $78.2\%$ on the validation data. With conjugate samples, our attentive action and context factorization method achieves a mean AP of $80.2\%$. This is better than the state-of-art performance \cite{m_Wang-Hoai-CVPR18} of $75.2\%$ mean AP. Figure \ref{fig:attmaps_voc} shows action and context attention maps on some action samples in the validation set.

\section{Conclusions}

We have presented a novel method for identifying action and context voxels of a video. Our method disentangles the action and the context components of a video with a novel attentional mechanism that can compute two spatiotemporal attention maps for action and context separately. Our method requires paired training data of action and conjugate samples of human actions, but this type of data can be collected easily. We have demonstrated the quantitative and qualitative benefits of using our method on the ActionThread, Hollywood2, HACS, and VOC Action datasets.

\iffalse

\begin{table}[t]
\begin{center}
\begin{tabular}{lcc}
\toprule
 & A & B \\
\midrule
 & & \\
 & & \\
\bottomrule
\end{tabular}
\end{center}
\caption{{\bf Title.} Description.}
\label{tab:name}
\end{table}

\begin{figure}[t]
\begin{center}
%\includegraphics[width=0.8\linewidth]{}
\end{center}
   \caption{{\bf Title.} Description.}
\label{fig:name}
\end{figure} 

\begin{equation}
\begin{split}
\phi &=\sum_{i}{w_i\phi_i}, \\
\textrm{with } w_i &= \frac{e^{-\alpha s_i}}{\sum_{j}{e^{-\alpha s_j}}}
\end{split}
\label{equ:name}
\end{equation}

\fi

{\small
\setlength{\bibsep}{0pt}
\bibliographystyle{abbrvnat}
\bibliography{shortstrings,pubs,egbib}
}

\end{document}

%% file: definitions.tex
\def\mA{\mathcal{A}}
\def\mB{\mathcal{B}}
\def\mC{\mathcal{C}}
\def\mD{\mathcal{D}}
\def\mE{\mathcal{E}}
\def\mF{\mathcal{F}}
\def\mG{\mathcal{G}}
\def\mH{\mathcal{H}}
\def\mI{\mathcal{I}}
\def\mJ{\mathcal{J}}
\def\mK{\mathcal{K}}
\def\mL{\mathcal{L}}
\def\mM{\mathcal{M}}
\def\mN{\mathcal{N}}
\def\mO{\mathcal{O}}
\def\mP{\mathcal{P}}
\def\mQ{\mathcal{Q}}
\def\mR{\mathcal{R}}
\def\mS{\mathcal{S}}
\def\mT{\mathcal{T}}
\def\mU{\mathcal{U}}
\def\mV{\mathcal{V}}
\def\mW{\mathcal{W}}
\def\mX{\mathcal{X}}
\def\mY{\mathcal{Y}}
\def\mZ{\mathcal{Z}}

\def\1n{\mathbf{1}_n}
\def\0{\mathbf{0}}
\def\1{\mathbf{1}}

\def\A{{\bf A}}
\def\B{{\bf B}}
\def\C{{\bf C}}
\def\D{{\bf D}}
\def\E{{\bf E}}
\def\F{{\bf F}}
\def\G{{\bf G}}
\def\H{{\bf H}}
\def\I{{\bf I}}
\def\J{{\bf J}}
\def\K{{\bf K}}
\def\L{{\bf L}}
\def\M{{\bf M}}
\def\N{{\bf N}}
\def\O{{\bf O}}
\def\P{{\bf P}}
\def\Q{{\bf Q}}
\def\R{{\bf R}}
\def\S{{\bf S}}
\def\T{{\bf T}}
\def\U{{\bf U}}
\def\V{{\bf V}}
\def\W{{\bf W}}
\def\X{{\bf X}}
\def\Y{{\bf Y}}
\def\Z{{\bf Z}}

\def\a{{\bf a}}
\def\b{{\bf b}}
\def\c{{\bf c}}
\def\d{{\bf d}}
\def\e{{\bf e}}
\def\f{{\bf f}}
\def\g{{\bf g}}
\def\h{{\bf h}}
\def\i{{\bf i}}
\def\j{{\bf j}}
\def\k{{\bf k}}
\def\l{{\bf l}}
\def\m{{\bf m}}
\def\n{{\bf n}}
\def\o{{\bf o}}
\def\p{{\bf p}}
\def\q{{\bf q}}
\def\r{{\bf r}}
\def\s{{\bf s}}
\def\t{{\bf t}}
\def\u{{\bf u}}
\def\v{{\bf v}}
\def\w{{\bf w}}
\def\x{{\bf x}}
\def\y{{\bf y}}
\def\z{{\bf z}}

\def\balpha{\mbox{\boldmath{$\alpha$}}}
\def\bbeta{\mbox{\boldmath{$\beta$}}}
\def\bdelta{\mbox{\boldmath{$\delta$}}}
\def\bgamma{\mbox{\boldmath{$\gamma$}}}
\def\blambda{\mbox{\boldmath{$\lambda$}}}
\def\bsigma{\mbox{\boldmath{$\sigma$}}}
\def\btheta{\mbox{\boldmath{$\theta$}}}
\def\bomega{\mbox{\boldmath{$\omega$}}}
\def\bxi{\mbox{\boldmath{$\xi$}}}
\def\bnu{\mbox{\boldmath{$\nu$}}}                                  
\def\bphi{\mbox{\boldmath{$\phi$}}}
\def\bmu{\mbox{\boldmath{$\mu$}}}

\def\bDelta{\mbox{\boldmath{$\Delta$}}}
\def\bOmega{\mbox{\boldmath{$\Omega$}}}
\def\bPhi{\mbox{\boldmath{$\Phi$}}}
\def\bLambda{\mbox{\boldmath{$\Lambda$}}}
\def\bSigma{\mbox{\boldmath{$\Sigma$}}}
\def\bGamma{\mbox{\boldmath{$\Gamma$}}}

\newcommand{\myminimum}[1]{\mathop{\textrm{minimum}}_{#1}}
\newcommand{\mymaximum}[1]{\mathop{\textrm{maximum}}_{#1}}    
\newcommand{\mymin}[1]{\mathop{\textrm{minimize}}_{#1}}
\newcommand{\mymax}[1]{\mathop{\textrm{maximize}}_{#1}}
\newcommand{\mymins}[1]{\mathop{\textrm{min.}}_{#1}}
\newcommand{\mymaxs}[1]{\mathop{\textrm{max.}}_{#1}}  
\newcommand{\myargmin}[1]{\mathop{\textrm{argmin}}_{#1}} 
\newcommand{\myargmax}[1]{\mathop{\textrm{argmax}}_{#1}} 
\newcommand{\myst}{\textrm{s.t. }}

\newcommand{\denselist}{\itemsep -1pt}
\newcommand{\sparselist}{\itemsep 1pt}

\definecolor{pink}{rgb}{0.9,0.5,0.5}
\definecolor{purple}{rgb}{0.5, 0.4, 0.8}   
\definecolor{gray}{rgb}{0.3, 0.3, 0.3}
\definecolor{mygreen}{rgb}{0.2, 0.6, 0.2}

\newcommand{\cyan}[1]{\textcolor{cyan}{#1}}
\newcommand{\red}[1]{\textcolor{red}{#1}}  
\newcommand{\blue}[1]{\textcolor{blue}{#1}}
\newcommand{\magenta}[1]{\textcolor{magenta}{#1}}
\newcommand{\pink}[1]{\textcolor{pink}{#1}}
\newcommand{\green}[1]{\textcolor{green}{#1}} 
\newcommand{\gray}[1]{\textcolor{gray}{#1}}    
\newcommand{\mygreen}[1]{\textcolor{mygreen}{#1}}    
\newcommand{\purple}[1]{\textcolor{purple}{#1}}       

\definecolor{greena}{rgb}{0.4, 0.5, 0.1}
\newcommand{\greena}[1]{\textcolor{greena}{#1}}

\definecolor{bluea}{rgb}{0, 0.4, 0.6}
\newcommand{\bluea}[1]{\textcolor{bluea}{#1}}
\definecolor{reda}{rgb}{0.6, 0.2, 0.1}
\newcommand{\reda}[1]{\textcolor{reda}{#1}}

\def\changemargin#1#2{\list{}{\rightmargin#2\leftmargin#1}\item[]}
\let\endchangemargin=\endlist
                                               
\newcommand{\cm}[1]{}

\newcommand{\mtodo}[1]{{\color{red}$\blacksquare$\textbf{[TODO: #1]}}}
\newcommand{\myheading}[1]{\vspace{1ex}\noindent \textbf{#1}}
\newcommand{\htimesw}[2]{\mbox{$#1$$\times$$#2$}}

% The following are useful for creating homework or exams

\newif\ifshowsolution
%\showsolutionfalse
\showsolutiontrue

\ifshowsolution  
\newcommand{\Comment}[1]{\paragraph{\bf $\bigstar $ COMMENT:} {\sf #1} \bigskip}
\newcommand{\Solution}[2]{\paragraph{\bf $\bigstar $ SOLUTION:} {\sf #2} }
\newcommand{\Mistake}[2]{\paragraph{\bf $\blacksquare$ COMMON MISTAKE #1:} {\sf #2} \bigskip}
\else
\newcommand{\Solution}[2]{\vspace{#1}}
\fi

\newcommand{\truefalse}{
\begin{enumerate}
	\item True
	\item False
\end{enumerate}
}

\newcommand{\yesno}{
\begin{enumerate}
	\item Yes
	\item No
\end{enumerate}
}